\RequirePackage{fix-cm}
\documentclass[twocolumn]{svjour3}

\smartqed  

\usepackage{amsmath,amssymb,amsfonts}
\usepackage{algorithmic}
\usepackage{graphicx}
\usepackage{textcomp}
\usepackage{algorithm}
\usepackage{natbib}

\newcommand{\argmin}{\mathop{\rm arg~min}\limits}

\begin{document}

\title{Constrained Mutual Convex Cone Method for Image Set Based Recognition
}

\author{
Naoya Sogi \and
Rui Zhu \and
Jing-Hao Xue \and
Kazuhiro Fukui 
}
\institute{
Naoya Sogi \at
University of Tsukuba, Japan\\
\email{sogi@cvlab.cs.tsukuba.ac.jp}
\and
Rui Zhu \at
City, University of London, United Kingdom \\
\email{rui.zhu@city.ac.uk}
\and
Jing-Hao Xue \at
University College London, United Kingdom\\
\email{jinghao.xue@ucl.ac.uk}
\and
Kazuhiro Fukui \at
University of Tsukuba, Japan\\
 \email{kfukui@cs.tsukuba.ac.jp}
}

\date{Received: date / Accepted: date}

\maketitle

\begin{abstract}
In this paper, we propose a method for image-set classification based on convex cone models.
Image set classification aims to classify a set of images, which were usually obtained from video frames or multi-view cameras, into a target object.
To accurately and stably classify a set, it is essential to represent structural information of the set accurately.
There are various representative image features, such as histogram based features, HLAC, and Convolutional Neural Network (CNN) features. We should note that most of them have non-negativity and thus can be effectively represented by a convex cone.
This leads us to introduce the convex cone representation to image-set classification.
To establish a convex cone based framework, we mathematically define multiple angles between two convex cones, and then define the geometric similarity between the cones using the angles. 
Moreover, to enhance the framework, we introduce a discriminant space that maximizes the between-class variance (gaps) and minimizes the within-class variance of the projected convex cones onto the discriminant space, similar to the Fisher discriminant analysis. 
Finally, the classification is performed based on the similarity between projected convex cones. 
The effectiveness of the proposed method is demonstrated experimentally by using five databases: CMU PIE dataset, ETH-80, CMU Motion of Body dataset, Youtube Celebrity dataset, and a private database of multi-view hand shapes.
\end{abstract}

\section{Introduction}
\sloppy
In this paper, we propose a method for image-set classification based on convex cone models, which can exactly represent the geometrical structure of an image set. In particular, we discuss the effectiveness of combining the proposed method and the convolutional neural network (CNN) features extracted from a high-level hidden layer of a learned CNN.

For the last decade, image set-based classification methods have gained substantial attention in various applications using multi-view images or videos, such as 3D object recognition and motion analysis. The essence of image set based classification is on how to effectively and low-costly measure the similarity between two image sets. To this end, several types of methods using different models have been proposed~\citep{cmsm,kmsm,komsm,kcmsm,gds,dcc,mmd,chisd,imageset1,mmdml,adnt,plrc,reconst,msm}.

In this paper, among the above various methods, we focus on subspace based methods, considering the compactness of a subspace model, simple geometrical relationship of class subspaces, and practical and efficient computation. In this type of method, a set of images is compactly modeled by a subspace in a high-dimensional vector space, where the subspace is generated by applying the Principal Component Analysis (PCA) to the image set without data centering. After converting each image set to a subspace, the similarity between two subspaces to be compared can be calculated by using the canonical angles between the subspaces~\citep{canangles2, canangles1}. Typical subspace-based methods are the mutual subspace method (MSM)~\citep{msm} and its extension, the constrained mutual subspace method (CMSM)~\citep{cmsm}.

Besides the above advantages, the validity of the subspace representation is also supported by the following physical characteristics: images of a convex object with Lambertian reflectance under various illumination conditions can be represented by a low-dimensional subspace, what is called an illumination subspace~\citep{illumcone1,illumcone2,illumcone3}. In other words, in object recognition, the subspace of an object can be stably generated from even few sample images under different illumination conditions.
Our representation by convex cone is an enhanced extension of the subspace representation.

Conventional subspace-based methods take a raw intensity vector or a hand-crafted feature as the input. Regarding more discriminant features, many recent studies have revealed that CNN features are effective inputs for various types of classifiers~\citep{cnnfeat1,cnnfeat_face,cnnfeat_sal,cnnfeat_maki}. 
Inspired by the successes in these studies, we expect that CNN features can also work as discriminant inputs for subspace based methods, such as MSM and CMSM. In this paper, we verify the effectiveness of CNN features for subspace based methods as the baseline. To the best of our knowledge, this paper is the first comprehensive report on the validity of the combination of MSM/CMSM and CNN features. 

\begin{figure}[!t]
\centering
\includegraphics[width=8.5cm]{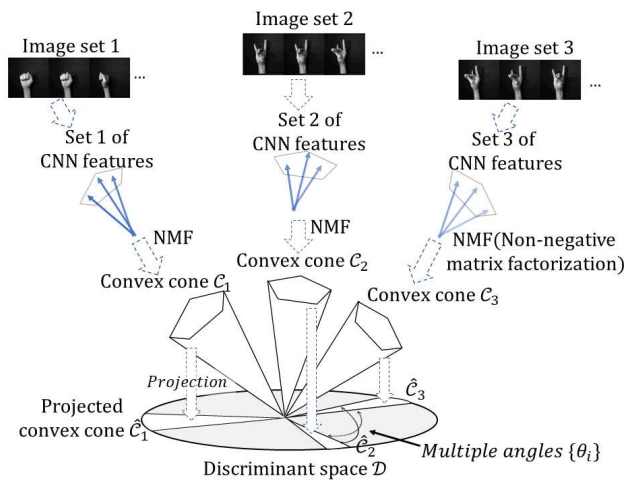}
\caption[Conceptual diagram of the proposed constrained mutual convex cone method (CMCM).]{Conceptual diagram of the proposed constrained mutual convex cone method (CMCM). First, a set of CNN features is extracted from an image set. Then, each set of CNN features is represented by a convex cone. After the convex cones are projected onto the discriminant space $\cal D$, the classification is performed by measuring similarity based on the angles $\{\theta_i\}$ between the two projected convex cones ${\hat{ \cal C}}_i$ and ${\hat{ \cal C}}_j$.}
\label{fig:abst}
\end{figure}

CNN feature vectors have only non-negative values when the rectified linear unit (ReLU)~\citep{relu} is used as an activation function. 
Although there are many types of features with non-negative constraint, in this paper, we focus on CNN features.
This characteristic of CNN features does not allow the combination of them with negative coefficients; accordingly, a set of CNN features forms a convex cone instead of a subspace in a high-dimensional vector space.

For example, it is well known that a set of front-facing images under various illumination conditions forms a convex cone, referred to as an illumination cone~\citep{illumcone1,illumcone2,illumcone3}. The illumination cone is a more strict representation than the illumination subspace mentioned above.
Several previous studies have demonstrated the advantages of convex cone representation compared with subspace representation~\citep{cone, kcone, mscd, hsicone}.
These advantages naturally motivated us to replace a subspace with a convex cone in models for a set of CNN features including the types of features with non-negative constraint. 

In this framework, it is necessary to consider how to calculate the geometric similarity between two convex cones. To this end, we define multiple angles between two convex cones by following the definition of the canonical angles~\citep{canangles1,canangles2} between two subspaces. 
Although the canonical angles between two subspaces can be analytically obtained from the orthonormal basis vectors of the two subspaces, the definition of angles between two convex cones is not trivial, as we need to consider the non-negative constraint.
In this paper, we define multiple angles between convex cones sequentially from the smallest to the largest by repeatedly applying the alternating least squares method~\citep{cone-cca}.
Then, the geometric similarity between two convex cones is defined based on the obtained angles.
We call the classification method using this similarity index the {\it mutual convex cone method} (MCM), corresponding to the mutual subspace method (MSM).

Moreover, to enhance the performance of the MCM, we introduce a discriminant space $\cal D$, which maximizes the between-class variance (gap) among convex cones projected onto the discriminant space and minimizes the within-class variance of the projected convex cones, similar to the Fisher discriminant analysis~\citep{fda}. 
The class separability can be increased by projecting the class of convex cones $\{{\cal C}_c\}$ onto the discriminant space $\cal D$, as shown in Fig.\ref{fig:abst}. As a result, the classification ability of MCM is enhanced, similar to that of the projection of class subspaces onto a generalized difference subspace (GDS) in CMSM~\citep{gds}. Finally, we perform the classification using the angles between the projected convex cones $\{{\hat{\cal C}}_c\}$.
We call this enhanced method the ``{\it constrained mutual convex cone method} (CMCM)," corresponding to the constrained MSM (CMSM).
This idea has been motivated by our previous preliminary work in~\citep{ijcnnpaper} and this paper shows more deep analysis with extensive and comprehensive experiments.

The main contributions of this paper are summarized as follows.
\begin{enumerate}
\item We verify the validity of the combination of MSM/ CMSM and CNN features, which has not yet been reported in the research fields of computer vision and machine learning.
\item To enhance the framework of the subspace based methods, we introduce a convex cone representation to accurately and compactly represent a set of features with non-negative constraint as typified by CNN features.
\item We introduce two novel mechanisms in our image set based classification: a) multiple angles between two convex cones to measure the similarity between the cones; and b) a discriminant space to increase the class separability among convex cones.
\item We propose two novel image set based classification methods, called MCM and CMCM, based on convex cone representation and the discriminant space.
\end{enumerate}

The paper is organized as follows. 
In Section~\ref{s:related}, we describe the algorithms of conventional methods, such as MSM and CMSM.
In Section~\ref{s:proposed}, we describe the details of the proposed method.
In Section~\ref{s:experiments}, we demonstrate the validity of the proposed method by visualization and classification experiments using four public datasets, i.e., CMU PIE~\citep{pie}, ETH-80~\citep{eth}, CMU Motion of Body \citep{mobo}, and Youtube Celebrity~\citep{ytc}, and a private database of multi-view hand shapes. 
Section~\ref{s:conclusion} concludes the paper.

\section{Related work}\label{s:related}
In this section, we first describe the algorithms for the MSM and CMSM, which are standard methods for image set classification. Then, we provide an overview of the concept of convex cones.

\subsection{Mutual subspace method based on canonical angles}
\begin{figure}[bt]
  \begin{center}
    \includegraphics[trim=0 0 0 0,clip,width=6cm]{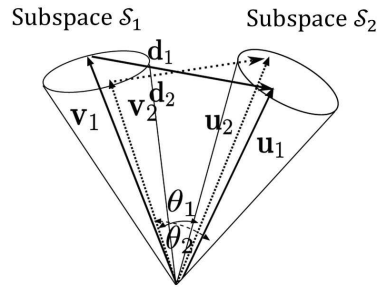}\vspace{2mm}\\
    \caption[Conceptual diagram of the canonical angles and canonical vectors.]{Conceptual diagram of the canonical angles and canonical vectors. The 1-st canonical vectors ${\bf u}_1,{\bf v}_1$ form the smallest angle $\theta_1$ between the subspaces. The 2-nd canonical vectors ${\bf u}_2,{\bf v}_2$ form the smallest angle $\theta_2$ in a direction orthogonal to $\theta_1$.}
    \label{dspace}
  \end{center}
\end{figure}

MSM is a classifier based on canonical angles between two subspaces, where each subspace represents an image set.

Given $N_1$-dimensional subspace ${\cal S}_1$ and $N_2$- dimensional subspace ${\cal S}_2$ in $d$-dimensional vector space, where $N_1 \leq N_2$, the canonical angles $\{0\leq \theta_1,\cdots,\theta_{N_1}\leq\frac{\pi}{2}\}$ between ${\cal S}_1$ and ${\cal S}_2$ are recursively defined as follows~\citep{canangles1,canangles2}:
\begin{eqnarray}
\cos{\theta_i}=\max_{{\bf u}\in {\cal S}_1}\max_{{\bf v}\in {\cal S}_2}{\bf u}^\mathrm{T}{\bf v}={\bf u}_i^\mathrm{T}{\bf v}_i, \\
s.t.\ \|{\bf u}_i\|_2=\|{\bf v}_i\|_2=1,{\bf u}_i^\mathrm{T}{\bf u}_j={\bf v}_i^\mathrm{T}{\bf v}_j=0,i\neq j, \nonumber
\end{eqnarray}
where ${\bf u}_i$ and ${\bf v}_i$ are the canonical vectors forming the $i$-th smallest canonical angle $\theta_i$ between ${\cal S}_1$ and ${\cal S}_2$. The $j$-th canonical angle $\theta_j$ is the smallest angle in the direction orthogonal to the canonical angles $\{\theta_k\}_{k=1}^{j-1}$ as shown in Fig.\ref{dspace}.

The canonical angles can be calculated from the orthogonal projection matrices onto subspaces ${\cal S}_1$ and ${\cal S}_2$.
Let $\{{\bf \Phi}_i\}_{i=1}^{N_1}$ be basis vectors of ${\cal S}_1$ and $\{{\bf \Psi}_i\}_{i=1}^{N_2}$ be basis vectors of ${\cal S}_2$. 
The projection matrices ${\bf{P}}_1$ and ${\bf{P}}_2$ are calculated as $\sum_{i=1}^{N_1} {{\bf{\Phi}}_i} {{\bf{\Phi}}_i}^\mathrm{T}$ and $\sum_{i=1}^{N_2} {{\bf{\Psi}}_i}{{\bf{\Psi}}_i}^\mathrm{T}$, respectively.  
$\cos^2 \theta_i$ is the $i$-th largest eigenvalue of ${\bf P}_1^\mathrm{T}{\bf P}_2$ or ${\bf P}_2^\mathrm{T}{\bf P}_1$.
Alternatively, the canonical angles can be easily obtained by applying the SVD to the orthonormal basis vectors of the subspaces.

\begin{figure}[!t]
\centering
\includegraphics[width=7.5cm]{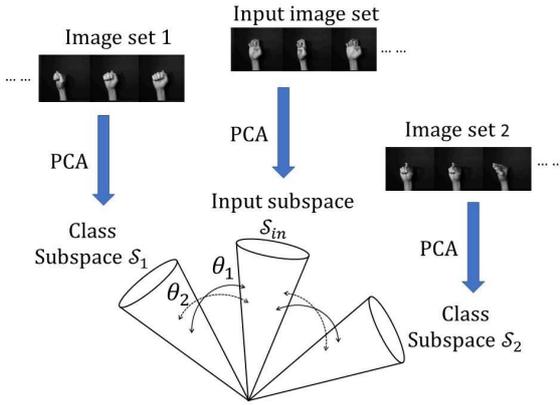}
\caption[Conceptual diagram of conventional MSM.]{Conceptual diagram of conventional MSM. Each image set is represented by a subspace, which is generated by applying the PCA to the set. In classification, the similarity between two subspaces is measured based on the canonical angles between them. An input subspace is assigned to the class of the subspace with the greatest similarity.}
\label{fig:msm}
\end{figure}

The geometric similarity between two subspaces ${\cal S}_1$ and ${\cal S}_2$ is defined by using the canonical angles as follows:
\begin{equation}
sim({\cal S}_1, {\cal S}_2)=\frac{1}{N_1}\sum_{i=1}^{N_1}{\cos^2 \theta_i}.
\end{equation}
In MSM, an input subspace ${\cal S}_{in}$ is classified by comparison with class subspaces $\{{\cal S}_c\}_{c=1}^C$ using this similarity as shown in Fig.\ref{fig:msm}.

\subsection{Constrained MSM}
\begin{figure}[!t]
\centering
\includegraphics[width=8.5cm]{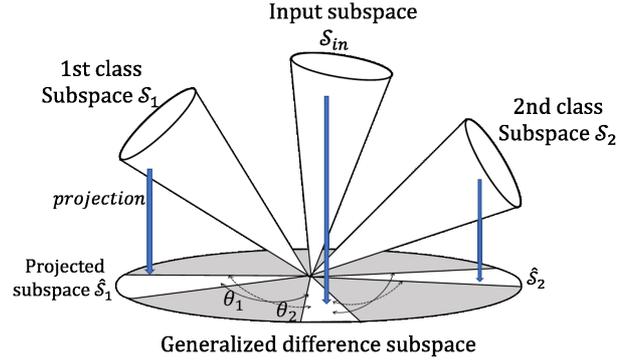}
\caption[Conceptual diagram of the constrained MSM (CMSM).]{Conceptual diagram of the constrained MSM (CMSM). By projecting class subspaces onto the generalized difference subspace, the separability between the classes is increased. By measuring the similarities among the projected subspaces using the canonical angles, the input subspace is assigned to either class 1 or 2.}
\label{fig:cmsm}
\end{figure}

The essence of the constrained MSM (CMSM) is the application of the MSM to a generalized difference subspace (GDS)~\citep{gds}, as shown in Fig.\ref{fig:cmsm}. GDS is designed to contain only difference components among subspaces $\{{\cal S}_c\}_{c=1}^C$. Thus, the projection of class subspaces onto GDS can increase the class separability among the class subspaces, substantially improving the classification ability of MSM \citep{gds}.

\subsection{Convex cone model}
\label{sec:cone}
In this subsection, we explain the definition of a convex cone and the projection of a vector onto a convex cone.
A convex cone $\cal{C}$ is defined by finite basis vectors $\{{\bf b}_i\}_{i=1}^r$ as follows:
\begin{equation}
{\cal C}=\{{\bf a} | {\bf a} = \sum_{i=1}^r{w_i{\bf b}_i}, w_i\geq 0\}.
\end{equation}
As indicated by this definition, the difference between the concepts of a subspace and a convex cone is whether there are non-negative constraints on the combination coefficients $w_i$ or not.

Given a set of feature vectors $\{{\bf f}_i\}_{i=1}^N\in\mathbb{R}^d$, the basis vectors $\{{\bf b}_i\}_{i=1}^r$ of a convex cone representing the  distribution of $\{{\bf f}_i\}$ can be obtained by non-negative matrix factorization (NMF)~\citep{nmf,nmfnnls}.
Let ${\bf F} = [{\bf f}_1 {\bf f}_2 \ldots{\bf f}_N]\in\mathbb{R}^{d\times N}$ and ${\bf B} = [{\bf b}_1 {\bf b}_2 \ldots {\bf b}_r]\in\mathbb{R}^{d\times r}$.
NMF generates the basis vectors ${\bf B}$ by solving the following optimization problem:
\begin{equation}
\argmin_{{\bf B},{\bf W}} \|{\bf F-BW}\|_F~~ s.t.~~({\bf B})_{i,j},({\bf W})_{i,j}\geq 0,
\end{equation}
where $\|\cdot\|_F$ denotes the Frobenius norm.
We use the alternating non-negativity-constrained least squares-based method~\citep{nmfnnls} to solve this problem. 

Although the basis vectors can be easily obtained by the NMF, the projection of a vector onto the convex cone is slightly complicated by the non-negative constraint on the coefficients.
In \citet{cone}, a vector ${\bf x}$ is projected onto the convex cone by applying the non-negative least squares method \citep{nnlsq} as follows:
\begin{equation}
\argmin_{\{w_i\}}{\|{\bf x} - \sum_{i=1}^r{w_i{\bf b}_i}\|_2}~~ s.t.~~ w_i\geq 0.
\label{eq:nnls}
\end{equation}
The projected vector ${\hat {\bf x}}$ is obtained as ${\hat {\bf x}} = \sum_{i=1}^r{w_i{\bf b}_i}$.

In the end, the angle $\theta$ between the convex cone and a vector ${\bf x}$ can be calculated as follows:
\begin{equation}
\cos\theta =\frac{{\bf x}^{\mathrm{T}}{\hat {\bf x}}}{\|{\bf x}\|_2\|{\hat {\bf x}}\|_2}.
\end{equation}

\section{Proposed method}\label{s:proposed}
In this section, we explain the algorithms in the MCM and CMCM, after establishing the definition of geometric similarity between two convex cones. 

\subsection{Geometric similarity between two convex cones}
\label{sec:angles}

We define the geometric similarity between two convex cones. To this end, we consider how to define multiple angles between two convex cones like canonical angles.
Two convex cones ${\cal C}_1$ and ${\cal C}_2$ are formed by basis vectors $\{{\bf b}^1_i\}_{i=1}^{N_1}\in \mathbb{R}^d$ and $\{{\bf b}^2_i\}_{i=1}^{N_2}\in \mathbb{R}^d$, respectively. Assume that $N_1 \leq N_2$ for convenience.
The angles between two convex cones cannot be obtained analytically like the canonical angles between two subspaces, as it is necessary to consider non-negative constraint. 
Alternatively, we find two vectors, ${\bf p} \in {\cal C}_1$ and ${\bf q} \in {\cal C}_2$, which are closest to each other. Then, we define the angle between the two convex cones as the angle formed by the two vectors. In this way, we sequentially define multiple angles from the smallest to the largest, in order.

\begin{figure}[!t]
\centering
\includegraphics[width=8cm]{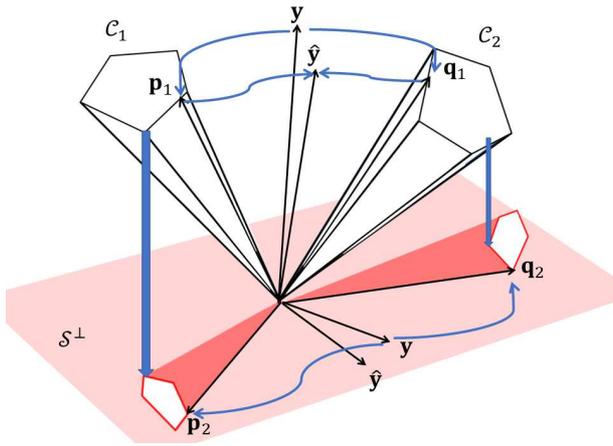}
\caption[Conceptual diagram of the procedure for searching pairs of vectors.]{Conceptual diagram of the procedure searching for pairs of vectors $\{{\bf p}_i,{\bf q}_i\}$. The first pair of ${\bf p}_1$ and ${\bf q}_1$ can be found by the alternating least squares method. The second pair of ${\bf p}_2$ and ${\bf q}_2$ is obtained by searching the orthogonal complement space ${\cal S}^{\perp}$ of ${\cal S}=$ Span$\{{\bf p}_1,{\bf q}_1\}$.}
\label{fig:near_vec}
\end{figure}

First, we search for a pair of $d$-dimensional vectors ${\bf p}_1\in {\cal C}_1$ and ${\bf q}_1\in {\cal C}_2$, which have the maximum correlation, using the alternating least squares method (ALS)~\citep{cone-cca}. The first angle $\theta_1$ is defined as the angle formed by ${\bf p}_1$ and ${\bf q}_1$. The pair of ${\bf p}_1$ and ${\bf q}_1$ can be found by using the following algorithm:

\vspace{2mm}
\noindent{\bf Algorithm to search for the pair ${\bf p}_1$ and ${\bf q}_1$}\\
Let ${\cal P}_1({\bf y})$ and ${\cal P}_2({\bf y})$ be the projections of a vector $\bf{y}$ onto ${\cal C}_1$ and ${\cal C}_2$, respectively. For the details of the projection, see Section~\ref{sec:cone}.
\begin{enumerate}
\item Randomly initialize ${\bf{y}}\in \mathbb{R}^d$.
\item ${\bf{p}}_1={\cal P}_1({\bf{y}}) / \|{\cal P}_1({\bf{y}})\|_2$. 
\item ${\bf{q}}_1={\cal P}_2({\bf{y}}) / \|{\cal P}_2({\bf{y}})\|_2$.
\item ${\hat {\bf y}} = ({\bf p}_1+{\bf q}_1) / 2$.
\item If $\|{\hat{\bf y}}-{\bf y}\|_2$ is sufficiently small, the procedure is completed. Otherwise, return to 2) setting ${\bf y}={\hat{\bf y}}$.
\item $\cos^2\theta_1={(\frac{{\bf p}_1^\mathrm{T}{\bf q}_1}{\|{\bf p}_1\|_2\|{\bf q}_1\|_2})^2}$.
\end{enumerate}
For the second angle $\theta_2$, we search for a pair of vectors ${\bf p}_2$ and ${\bf q}_2$ with the maximum correlation, but with the minimum correlation with ${\bf p}_1$ and ${\bf q}_1$. Such a pair can be found by applying ALS to the projected convex cones ${\cal C}_1$ and ${\cal C}_2$ on the orthogonal complement space ${\cal S}^{\perp}$ of the subspace ${\cal S}$ spanned by the vectors ${\bf p}_1$ and ${\bf q}_1$ as shown in Fig.\ref{fig:near_vec}. Then $\theta_2$ is formed by ${\bf p}_2$ and ${\bf q}_2$. In this way, we can obtain all of the pairs of vectors ${\bf p}_i, {\bf q}_i$ forming the $i$-th angle $\theta_i$, $i=1,\dots,N_1$.

With the resulting angles $\{\theta_i\}_{i=1}^{N_1}$, we define the geometrical similarity $sim$ between two convex cones ${\cal C}_1$ and ${\cal C}_2$ as follows:
\begin{equation}
sim({\cal C}_1,{\cal C}_2) = \frac{1}{N_1}\sum_{i=1}^{N_1}{\cos^2\theta_i}.
\label{eq:sim_cone}
\end{equation}

\subsection{Mutual convex cone method}
\begin{figure}[!t]
\centering
\includegraphics[width=8.5cm]{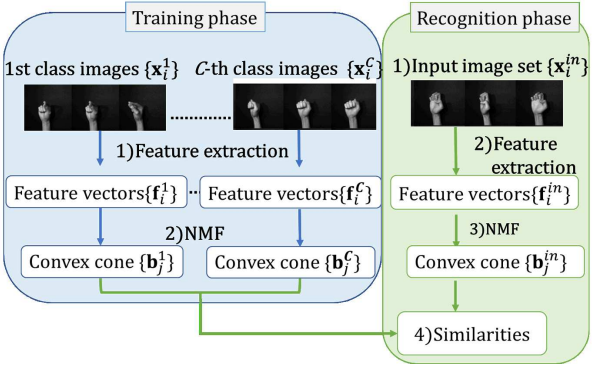}
\caption{Process flow of the proposed mutual convex cone method (MCM), which consists of a training phase and a recognition phase.}
\label{fig:flow_mcm}
\end{figure}

The mutual convex cone method (MCM) classifies an input convex cone based on the similarities defined by Eq.(\ref{eq:sim_cone}) between the input and the class convex cones.
MCM consists of two phases, a training phase and a recognition phase, as summarized in Fig.\ref{fig:flow_mcm}. 

Given $C$ class sets with $L$ images $\{{\bf x}^c_i\}_{i=1}^L$.

\vspace{2mm}
\noindent{\bf Training Phase}
\begin{enumerate}
\item Feature vectors $\{{\bf f}_i^c\}$ are extracted from the images $\{{\bf x}_i^c\}$ of class $c$.
\item The basis vectors of class-$c$ convex cone, $\{{\bf b}_j^c\}$, are generated by applying NMF to the set of feature vectors $\{{\bf f}_i^c\}$.
\item $\{{\bf b}_{j}^c\}$ are registered as the reference convex cone of class $c$.
\item The above process is conducted for all $C$ classes.
\end{enumerate}

\vspace{2mm}
\noindent{\bf Recognition Phase}
\begin{enumerate}
\item A set of images $\{{\bf x}_i^{in}\}$ is input.
\item Feature vectors $\{{\bf f}_i^{in}\}$ are extracted from the images $\{{\bf x}_i^{in}\}$.
\item The basis vectors of the input convex cone, $\{{\bf b}_j^{in}\}$, are generated by applying NMF to the input set of feature vectors.
\item The input image set $\{{\bf x}_i^{in}\}$ is classified based on the similarity (Eq.(\ref{eq:sim_cone})) between the input convex cone $\{{\bf b}_j^{in}\}$ and the $c$-th class reference convex cone $\{{\bf b}_j^c\}$.
\end{enumerate}

\subsection{Generation of discriminant space}
To enhance the performance of the mutual convex cone method, we introduce a discriminant space $\cal D$, which maximizes the between-class variance $\bf S_b$ and minimizes the within-class variance $\bf S_w$ for the convex cones projected on $\cal D$, similarly to the Fisher discriminant analysis (FDA). In our method, the within-class variance $\bf S_w$ is calculated from basis vectors of convex cones, and the between-class variance $\bf S_b$ is calculated from gaps among convex cones for effectively utilizing the information formed by convex cones.

We define these gaps as follows.
Let ${\cal C}_c$ be the $c$-th class convex cone with basis vectors $\{{\bf b}_i^c\}_{i=1}^{N_c}$, ${\cal P}_c$ be the projection operation of a vector onto ${\cal C}_c$ defined by Eq.(\ref{eq:nnls}), and $C$ be the number of the classes. 
We consider $C$ vectors $\{{\bf p}_1^c\}$, $c=1,2,\ldots,C$, such that the sum of the correlation $\sum_{c{\neq}{c'}}{({\bf p}_1^c)^{\mathrm{T}}}{{\bf p}_1^{c'}/(\|{\bf p}_1^{c}\|_2\|{\bf p}_1^{c'}\|_2)}$ is maximum. Such a set of vectors can be obtained by using the following algorithm. This algorithm is almost the same as the generalized canonical correlation analysis~\citep{mcca1,mcca2}, except that the non-negative least squares (LS) method is used instead of the standard LS method.

\vspace{2mm}
\noindent{\bf Procedure to search for a set of first vectors $\{{\bf p}_1^c\}_{c=1}^C$}
\begin{enumerate}
\item Randomly initialize ${\bf y}_1$.
\item Project ${\bf y}_1$ onto each convex cone, and then normalize the projection as ${\bf p}_1^c={\cal P}_c({\bf y}_1)/\|{\cal P}_c({\bf y}_1)\|_2$.
\item ${\hat {\bf y}_1}=\sum_{c=1}^C{{\bf p}_1^c}/C$.
\item If $\|{\bf y}_1-{\hat{\bf y}}_1\|_2$ is sufficiently small, the procedure is completed. Otherwise, return to 2) setting ${\bf y}_1={\hat{\bf y}}_1$.
\end{enumerate}

Next, we search for a set of second vectors $\{{\bf p}_2^c\}$ with the maximum sum of the correlations under the constraint condition that they have the minimum correlation with the previously found $\{{\bf p}_1^c\}$.
The second vectors $\{{\bf p}_2^c\}$ can be obtained by applying the above procedure to the convex cones projected onto the orthogonal complement space of the vector ${\bf y}_1$. 
In the following, a set of the $j$-th vectors $\{{\bf p}_j^c\}$ can be sequentially obtained by applying the same procedure to the convex cones projected onto the orthogonal complement space of $\{{\bf y}_k\}_{k=1}^{j-1}$. In this way, we finally obtain the sets of $\{{\bf p}_j^c\}$.
With the sets of $\{{\bf p}_j^c\}$, we define a difference vector $\{{\bf d}_{j}^{c_1c_2}\}$ as follows:
\begin{equation}
{\bf d}_{j}^{c_1c_2}= {\bf p}_j^{c_1}-{\bf p}_j^{c_2}.
\label{eq:gap}
\end{equation}
Considering that each difference vector represents the gap between the two convex cones, we define $\bf S_b$ using these vectors as follows:
\begin{equation}
{\bf S_b}=\sum_{j=1}^{N_g}{\sum_{c_1=1}^{C-1}{\sum_{c_2=c_1+1}^{C}{{\bf d}_{j}^{c_1c_2}({\bf d}_{j}^{c_1c_2})^\mathrm{T}}}},
\end{equation}
where $N_g$ can be set from 1 to $\min(\{N_c\})$.

Next, we define the within-class variance ${\bf S_w}$ using the basis vectors $\{{\bf b}_i^c\}$ for all classes of convex cones as follows:
\begin{equation}
{\bf S_w} = \sum_{c=1}^C{\sum_{i=1}^{N_c}{({\bf b}_i^c-{\bf \mu}_c)({\bf b}_i^c-{\bf \mu}_c)^\mathrm{T}}},
\end{equation}
where ${\bf \mu}_c=\sum_{i=1}^{N_c}{{\bf b}_i^c}/{N_c}$.
Finally, the $N_d$-dimensional discriminant space $\cal D$ is spanned by $N_d$ eigenvectors $\{\phi_i\}_{i=1}^{N_d}$ corresponding to the $N_d$ largest eigenvalues $\{\gamma_i\}_{i=1}^{N_d}$ of the following eigenvalue problem:
\begin{equation}
{\bf S_b}{\bf \phi}_i=\gamma_i{\bf S_w}{\bf \phi}_i.
\label{eigenpr}
\end{equation}
\label{sec:gdsc}

\subsection{Constrained mutual convex cone method}
\begin{figure}[!t]
\centering
\includegraphics[width=8cm]{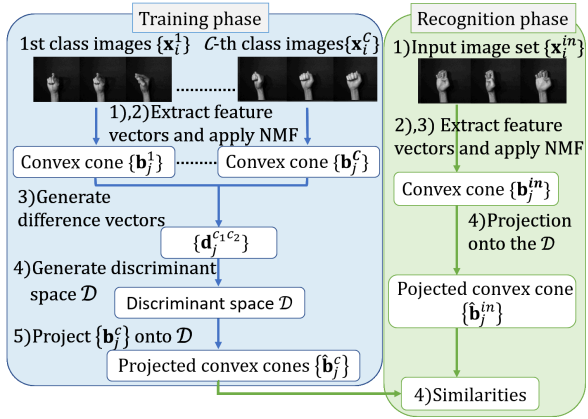}
\caption[Process flow of the proposed constrained MCM (CMCM).]{Process flow of the proposed constrained MCM (CMCM). CMCM is an enhanced version of MCM with the projection of class convex cones onto the discriminant space $\cal D$.}
\label{fig:flow_cmcm}
\end{figure}
We construct the constrained MCM (CMCM) by incorporating the projection onto the discriminant space $\cal D$ into the MCM. 
CMCM consists of a training phase and a recognition phase, as shown in Fig.\ref{fig:flow_cmcm}.
In the following, we explain each phase for the case in which $C$ classes have $L$ images $\{{\bf x}^c_i\}_{i=1}^L$ each.

\newpage
\noindent{\bf Training Phase}
\begin{enumerate}
\item Feature vectors $\{{\bf f}_i^c\}$ are extracted from the images $\{{\bf x}_i^c\}$.
\item The basis vectors of the $c$-th class convex cone, $\{{\bf b}_j^c\}$, are generated by applying NMF to each class set of feature vectors.
\item Sets of difference vectors $\{{\bf d}_{j}^{c_1c_2}\}$ are generated according to the procedure described in section \ref{sec:gdsc}. 
\item The discriminant space ${\cal D}$ is generated by solving Eq.(\ref{eigenpr}) using $\{{\bf b}_j^c\}$ and $\{{\bf d}_{j}^{c_1c_2}\}$.
\item The basis vectors $\{{\bf b}_j^c\}$ are projected onto the discriminant space ${\cal D}$ and then the lengths of the projected basis vectors are normalized to 1.
A set of these  basis vectors $\{{\hat{\bf b}}_j^c\}$ forms the projected convex cone.
\item $\{{\hat{\bf b}}_{j}^c\}$ are registered as the reference convex cones of class $c$.
\end{enumerate}

\vspace{2mm}
\noindent{\bf Recognition Phase}
\begin{enumerate}
\item A set of images $\{{\bf x}_i^{in}\}$ is input.
\item Feature vectors $\{{\bf f}_i^{in}\}$ are extracted from the images $\{{\bf x}_i^{in}\}$. 
\item The basis vectors of a convex cone, $\{{\bf b}_j^{in}\}$, are generated by applying NMF to the set of feature vectors.
\item The basis vectors $\{{\bf b}_j^{in}\}$ are projected onto the discriminant space ${\cal D}$ and then the lengths of the projected basis vectors are normalized to 1. The normalized projections are represented by $\{{\hat {\bf b}}_j^{in}\}$.
\item The input set $\{{\bf x}_i^{in}\}$ is classified based on the similarity (Eq.(\ref{eq:sim_cone})) between the input convex cone $\{{\hat {\bf b}}_j^{in}\}$ and each class reference convex cone $\{{\hat {\bf b}}_j^c\}$.
\end{enumerate}

\section{Evaluation experiments}\label{s:experiments}
In this section, we demonstrate the effectiveness of the proposed methods through four experiments.
The first experiment uses the ETH-80 dataset to verify the effectiveness of using multiple angles between convex cones as the similarity between them.
The second experiment analyzes the attribute of difference vectors between two convex cones by visualizing the difference vectors as images.
The third experiment evaluates the classification performance of the proposed methods using the three datasets, 1) ETH-80~\citep{eth}, 2) CMU Motion of Body (CMU MoBo)~\citep{mobo}, and 3) YouTube Celebrities (YTC)~\citep{ytc}, with a large number of training samples.
The fourth experiment demonstrates the robustness of the proposed methods against the small sample sizes (SSS) problem, considering the situation in which only few training samples are available for learning. 
In this experiment, we use the multi-view hand shape dataset \citep{hand}

\subsection{Effectiveness of using multiple angles}
\begin{figure}[!t]
\centering
\includegraphics[width=8.5cm]{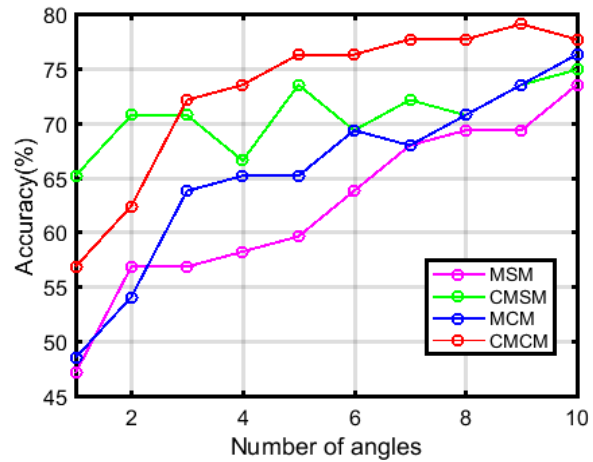}
\caption{Results of classification experiment. The vertical axis denotes accuracy, and the horizontal axis denotes the number of angles used for calculating the similarity.}
\label{fig:n_angle}
\end{figure}
In this experiment, we verify the effectiveness of using multiple angles for calculating the similarity between convex cones, through a classification experiment using the ETH-80 dataset.
The ETH-80 dataset consists of object images in eight different categories, captured from 41 viewpoints.
Each category has ten kinds of object.
One object randomly sampled from each category set was used for training, and the remaining nine objects were used for test.
As an input image set, we used 41 multi-view images for each object.
We used images scaled to 32 $\times$ 32 pixels and converted to grayscale.
Vectorized features of the grayscale images were used as input, i.e. the dimension of the feature vector is 1024.

We evaluated the classification performance of mutual convex cone method (MCM) and constrained MCM (CMCM), while varying the number of angles used for calculating the similarity.
As baselines, the mutual subspace method (MSM) and constrained MSM (CMSM) were also evaluated.
Dimensions of reference subspaces and convex cones were set to 20, and dimensions of input subspaces and convex cones were set to 10.

Fig.\ref{fig:n_angle} shows the accuracy changes of the different methods against the number of angles.
The horizontal axis denotes the number of angles used for calculating the similarity.
We can confirm that the accuracy of MCM and CMCM increases, as the number of angles increases.
This result shows clearly the importance of comparing the whole structures of convex cones by using multiple angles rather than using only the minimum angle for accurate classification.

In case of using one or two angles, the accuracy of CMCM is less than CMSM.
However, with an increase in the numbers of angles, CMCM outperforms the methods MSM and CMSM that are based on subspace representation. This indicates that using multiple angles is required to compare the structures of two convex cones.

\subsection{Validity of difference vectors between convex cones}
\begin{figure}[!t]
\centering
\includegraphics[width=8.5cm]{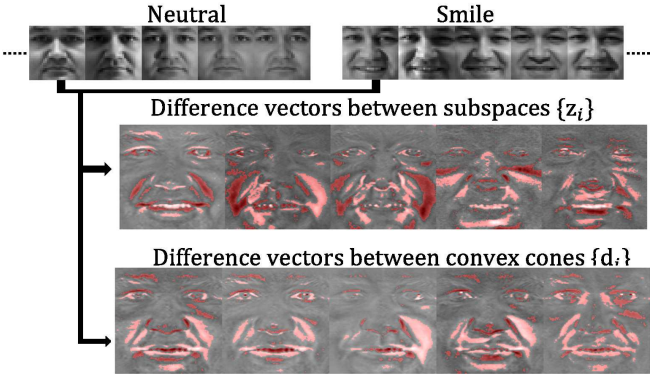}
\caption{Results of visualizing the difference vectors between two convex cones and difference vectors between the subspaces of neutral and simile. The parts with values larger than the threshold, which is automatically decided by Otsu's binarization~\citep{otsu_th}, in the difference vectors are emphasized in red.}
\label{fig:vis_gap}
\end{figure}

In this experiment, we demonstrate the validity of difference vectors, $\{{\bf{d}}\}$, between convex cones through the visualization of $\{{\bf{d}}\}$ on two sets of facial expressions, neutral and smile. They were extracted from the CMU PIE dataset \citep{pie}.
Each set has 20 front face images taken under various illumination conditions.

After representing the two sets of raw images as convex cones, we generated the difference vectors $\{{\bf{d}}_i\}$ between the two convex cones according to Eq.(\ref{eq:gap}).
For comparison, we also calculated the difference vectors $\{{\bf{z}}_i\}$ between the canonical vectors of two subspaces of the two sets.
We set the number of basis vectors of each convex cone to 5 and the dimension of each subspace to 5.

\begin{figure}[!t]
\centering
\includegraphics[width=5cm]{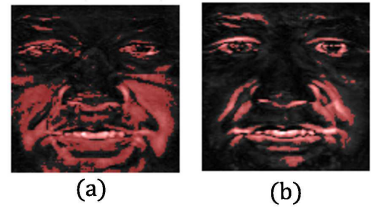}
\caption{Mean images of absolute value images of the difference vectors $\{{\bf d}_i\}_{i=1}^5$ between convex cones and the difference vectors $\{{\bf z}_i\}_{i=1}^5$ between subspaces. (a) $\sum_{i=1}^5{|{\bf z}_i|}/5$. (b) $\sum_{i=1}^5{|{\bf d}_i|}/5$. The parts with the values larger than the threshold, which is automatically decided by Otsu's binarization \citep{otsu_th}, in the difference vectors are emphasized in red.}
\label{fig:mean_abs}
\end{figure}

Fig.\ref{fig:vis_gap} shows the visualizations of $\{{\bf d}_i\}_{i=1}^5$ and $\{{\bf z}_i\}_{i=1}^5$.
We can see that both sets of the difference vectors can emphasize regions around smile lines and eyes. These regions can move largely in comparison with other regions when changing from neutral face expression to smile.
However, the resolutions in variation captured by them are a bit different. To take a closer look at this difference, we calculated mean images of the absolute values of the difference vectors, by $\sum_{i=1}^5{|{\bf d}_i|}/5$ and $\sum_{i=1}^5{|{\bf z}_i|}/5$, as shown in Fig.\ref{fig:mean_abs}.
The difference vectors, $\{{\bf{z}}_i\}$, between the subspaces capture roughly difference on the whole face.
On the other hand, the difference vectors, $\{{\bf{d}}_i\}$, between convex cones capture clearly fine difference on smile lines and around eyes.

\begin{figure}[!t]
\centering
\includegraphics[width=8.5cm]{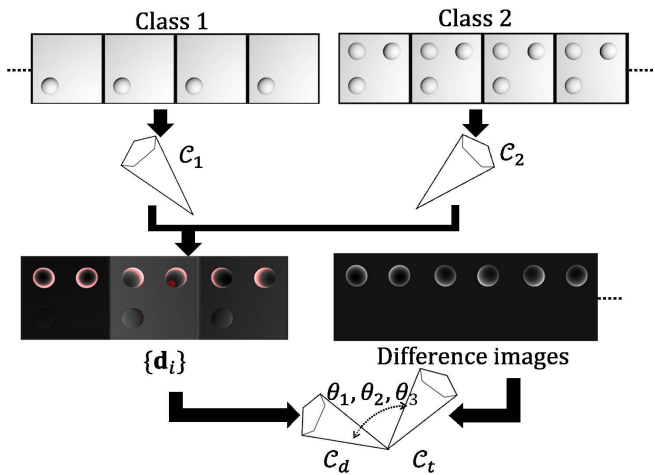}
\caption{Results of the experiment using synthesized data. After generating convex cones ${\cal C}_1, {\cal C}_2$ for each set, we calculated difference vectors $\{{\bf d}_i\}$ between ${{\cal C}_1}$ and ${{\cal C}_2}$. Then, we evaluated cosine similarities between two convex cones ${\cal C}_d$, ${\cal C}_t$, which are spanned by $\{{\bf d}_i\}$ and the difference images between pairs of original images, respectively.}
\label{fig:gap_synth}
\end{figure}

Besides, to verify how much a set of difference vectors between two convex cones captures the difference in the structure of them, we conducted a comparison experiment using two synthetic convex cones ${\cal{C}}_1$ and ${\cal{C}}_2$, which are shown in Fig.\ref{fig:gap_synth}. The convex cones are spanned by three basis vectors, which were generated by applying NMF to a set of images of two different objects synthesized under 100 illumination conditions. We calculated the difference vectors $\{{\bf{d}}_i\}$ between ${\cal{C}}_1$ and ${\cal{C}}_2$. Let the convex cone spanned by $\{{\bf{d}}_i\}$ be convex cone ${\cal{C}}_d$. Note that the $\{{\bf{d}}_i\}$ are not orthogonal to each other, so that they span a convex cone. 
Besides ${\cal{C}}_d$, we generated a convex cone ${\cal{C}}_t$, which is spanned by three basis vectors obtained by applying NMF to a set of difference image vectors between pairs of object images of classes 1 and 2. According to our definition, we expect that ${\cal{C}}_d$ can have a high correlation with ${\cal{C}}_t$. 
In fact, the first three cosine similarities between ${\cal{C}}_d$ and ${\cal{C}}_t$ are 0.9104, 0.8478, and 0.5426 , respectively.
The high correlations support that a set of the difference vectors, namely the convex cone spanned by them, captures effectively the structural difference between the convex cones.

\subsection{Comparison of classification performance with conventional methods}
In this subsection, we evaluate the classification performance of the proposed methods compared with various conventional methods using three public datasets.
In the following, details of each dataset and experimental protocols are described.
After that, experiment results are shown.

\subsubsection{ETH-80 dataset}
The ETH-80 dataset consists of eight different categories, captured from 41 viewpoints.
Each category has ten kinds of object.
Five objects randomly sampled from each category were used for training, and the remaining objects were used for testing.
As an input image set, we used 41 multi-view images for each object.
To conduct a consistent experiment with previous works, we used images scaled to 32 $\times$ 32 pixels~\citep{reconst, adnt}.
We evaluated the classification performance of each method in terms of the average accuracy of ten trials using randomly divided datasets.

For MSM and CMSM, the dimensions of class subspaces, input subspaces, and GDS were set to 50, 30, and 395, respectively.
For MCM and CMCM, the numbers of the basis vectors of class and input convex cones were set to 50 and 30, respectively. The dimension $N_d$ of the discriminant space $\cal D$ was set to 450.
We determined these dimensionalities by cross-validation using the training data.

In this experiment, we used CNN features as feature vectors.
To obtain CNN features under our experimental setting, we modified the original ResNet-50~\citep{resnet} trained by the ImageNet database~\citep{imagenet} slightly for our experimental conditions. First, we replaced the final 1000-way fully connected (FC) layer of the original ResNet-50 with a 1024-way FC layer and applied the ReLU function. Then, we added a $\it class~number$-way FC layer with softmax behind the previous 1024-way FC layer.

Moreover, to extract more effective CNN features from our modified ResNet, we fine-tuned our ResNet using the learning set. A CNN feature vector was extracted from the 1024-way FC layer every time an image was input into our ResNet. As a result, the dimensionality $d$ of a CNN feature vector was 1024. 

In our fine-tuned CNN, an input image set was classified based on the average value of the output conviction degrees for each class from the last FC layer with softmax.
In this section, we refer to this method as ``softmax".

\subsubsection{CMU MoBo dataset}
The CMU Mobo dataset~\citep{mobo} consists of 25 people videos walking on a treadmill.
Although the original purpose of this dataset was to research on human gait analysis~\citep{mobo}, in this experiment we conducted image set based face classification following previous works~\citep{reconst, adnt, chisd, mmd}.

The face images were detected by the Viola and Jones detection algorithm~\citep{violajones} from video frames.
Detected face images were reshaped to 40 $\times$ 40 pixels and converted to grayscale.
Face images extracted from one video was considered as an image set.

The dataset contains four walking patterns (videos) of each person, except for one person.
We used videos of 24 people with all walking patterns.
One video randomly sampled from each person was used for training, and the remaining three videos were used for testing.
We repeated the evaluation ten times with different random selections.

For MSM and CMSM, the dimensions of class subspaces, input subspaces, and GDS were set to 50, 50, and 1000, respectively.
For MCM and CMCM, the numbers of the basis vectors of class and input convex cones were set to 50 and 30, respectively. The dimension $N_d$ of the discriminant space $\cal D$ was set to 1000.
We determined these dimensionalities by cross-validation using the training data.
CNN features were extracted from the fine-tuned ResNet under this experimental setting, according to the same procedure used in the previous experiments.
\begin{table*}[t]
\begin{center}
\caption{Experimental results (recognition rate (\%), standard deviation) for the three public datasets.}
\begin{tabular}{|c||c|c|c|}
\hline
~ & ETH-80 & CMU Mobo & YTC  \\ \hline \hline
DCC\citep{dcc} & 91.75$\pm$3.74 & 88.89$\pm$2.45 & 51.42$\pm$4.95 \\\hline
MMD\citep{mmd} & 77.50$\pm$5.00 & 92.50$\pm$2.87 & 54.04$\pm$3.69 \\\hline
CHISD\citep{chisd} & 79.53$\pm$5.32 & 96.52$\pm$1.18 & 60.42$\pm$5.95 \\\hline
MMDML\citep{mmdml} & 94.5$\pm$3.5 & 97.8$\pm$1.0 & - \\\hline
ADNT\citep{adnt} & 98.12$\pm$1.69 & 97.92$\pm$0.73 & {\bf 71.35$\pm$4.83} \\ \hline
PLRC\citep{plrc} & 87.72$\pm$5.67 & 93.74$\pm$4.3 & 61.28$\pm$6.37 \\ \hline
Reconstruct Model \citep{reconst} & 94.75$\pm$4.32 & 98.33$\pm$1.27 & 66.45$\pm$5.07\\\hline
softmax & 96.50$\pm$2.29 & 98.61$\pm$1.52 & 64.18$\pm$2.20 \\ \hline \hline
			 
CNN feature + MSM  & 99.50$\pm$1.05 & 99.17$\pm$0.97 & 64.26$\pm$2.89 \\ \hline
CNN feature + CMSM & 99.50$\pm$1.05 & {\bf 99.58$\pm$0.67} & 66.45$\pm$2.36 \\ \hline  
CNN feature + MCM  & 99.50$\pm$1.05 & 98.75$\pm$1.22 & 64.11$\pm$2.68 \\ \hline
CNN feature + CMCM & {\bf 99.75$\pm$0.79} & {\bf 99.58$\pm$0.67} & 66.74$\pm$2.12 \\\hline
\end{tabular}
\label{tab:exp1}
\end{center}
\end{table*}

\subsubsection{YTC dataset}
The YTC dataset~\citep{ytc} contains 1910 videos of 47 people.
Similarly to \citep{reconst}, as an image set, we used a set of face images extracted from a video by the Incremental Learning Tracker~\citep{ilt}.
All the extracted face images were scaled to 30 $\times$ 30 pixels and converted to grayscale.
Three videos per each person were randomly selected as training data, and six videos per each person were randomly selected as test data.
We conducted five-fold cross-validation according to the above procedure.

For MSM and CMSM, the dimensions of class subspaces, input subspaces, and GDS were set to 70, 10, and 824, respectively.
For MCM and CMCM, the numbers of the basis vectors of class and input convex cones were set to 50 and 40, respectively. The dimension $N_d$ of the discriminant space $\cal D$ was set to 1000.
We determined these dimensionalities by cross-validation using the training data.
CNN features were extracted from the fine-tuned ResNet under this experimental setting, according to the same procedure used in the previous experiments.

\subsubsection{Results and discussion}
Table \ref{tab:exp1} shows the classification results of the proposed methods and various conventional methods, including several Deep Neural Networks based methods.
First of all, we can see that the subspace-based methods and the proposed MCM/CMCM achieve comparative or better performances than that of the conventional methods in all the datasets. In particular, it is notable that the proposed methods achieve competitive results with more complex methods using deep learning, such as softmax, MMDML and ADNT. Especially, in ETH-80 and Mobo, they show very high recognition rates against these deep learning based methods.
The conventional methods do not explicitly consider the structure information of an image set. In contrast, the proposed methods extract effectively the detailed structure information through the convex cone representation. This difference in the classification mechanism leads to the advantage of our methods.

\begin{figure}[!t]
\centering
\includegraphics[width=8cm]{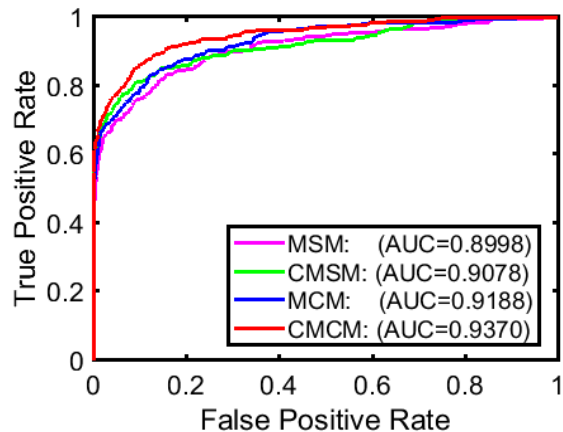} \\
\caption{ROC curves of subspace and convex cone based methods for the YTC dataset.}
\label{fig:ytc_roc}
\end{figure}

CMCM outperformed MCM in all the cases.
This indicates that projecting onto the discriminant space can capture useful geometrical information to increase the class separability among the class convex cones, as we expected. 
CMSM also improves the performance of MSM. However, the improvement degree by CMCM is larger than that of CMSM. This implies that the discriminant space works better with convex cone representation to enhance the class separability among class cones.

The results on ETH-80 and Mobo show clearly the effectiveness of both of cone and subspace based methods against the conventional methods. However, it may be difficult to argue the advantage of CMCM against CMSM, since they both realized almost 100$\%$ recognition rate with near zero EERs. The databases seemed to be relatively easy for both types of methods to classify.

On the other hand, the YTC is difficult for all the methods, so that we can find apparent difference between the recognition rates of both. To visually confirm this advantage, we calculated the receiver operating characteristic (ROC) curves of four subspace and cone based methods, as shown in Fig.\ref{fig:ytc_roc}. The ROC curves indicate clearly the strength of CMCM against CMSM.
This superiority is also supported by the average the area under the curve (AUC) as follows:~ CMSM and CMCM are 0.9002
and 0.9341
respectively.  


\subsection{Robustness against limited training data}
A major issue with deep neural networks is the requirement of a large number of training samples to learn the networks accurately. Therefore, the robustness against small sample size (SSS) is a necessary characteristic for effective methods using CNN features in practice.
In this experiment, we evaluated the robustness of the different methods against SSS using our private multi-view hand shape dataset~\citep{hand}.

\subsubsection{Experimental protocol}
\begin{figure}[!t]
\centering
\includegraphics[width=8cm]{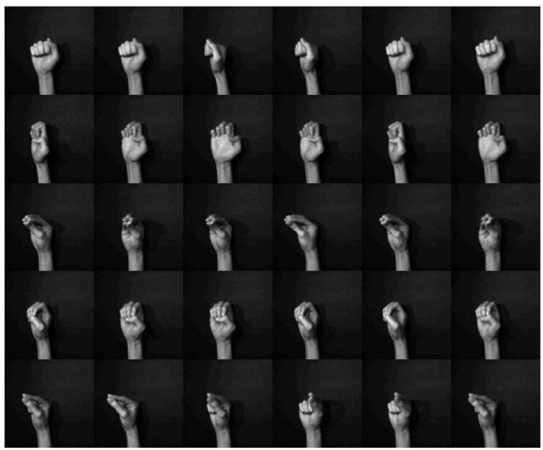}
\caption[Sample images of the multi-view hand shape dataset.]{Sample images of the multi-view hand shape dataset used in the experiments. Each row shows a hand shape from various viewpoints.}
\label{fig:dataset}
\end{figure}

The multi-view hand shape dataset consists of 30 classes of hand shapes.
Each class data was collected from 100 subjects at a speed of 1 fps for 4 s using a multi-camera system equipped with seven synchronized cameras at intervals of 10 degrees.
During data collection, the subjects were asked to rotate their hands at a constant speed to increase the number of viewpoints. 
Figure \ref{fig:dataset} shows several sample images in the dataset.
The total number of images collected was 84000 (= 30 classes$\times$4 frames$\times$7 cameras $\times$100 subjects).

We randomly divided the subjects into two sets. 
One set was used for training, and the other was used for testing.
We evaluated the performances of the methods by setting the numbers of subjects used for training to 1, 2, 3, 4, 5, 10, and 15.
In each case, the total number of training images was 30 classes$\times$7 cameras$\times$4 frames$\times$ $N$ subjects, ($N=1, 2, 3, 4, 5, 10, 15$).
We set the number of subjects used for testing to 50. As an input image set, we used 28 (=7 cameras $\times$4 frames) images of a subject. 
Thus, the total number of convex cones for testing was 1500 (=30 classes$\times$50 subjects).

To extract CNN features from the images, we used the fine-tuned ResNet by using the training images under the experimental conditions.

\begin{table}[t]
\begin{center}
\caption{Change in the accuracies ($\%$) against the number of training subjects.}
\begin{tabular}{|c||c|c|c|c|c|}
\hline
$N$ & softmax & MSM & CMSM & MCM & CMCM  \\ \hline \hline
1 & 36.07 & 62.27 & 65.87 & 63.07 & 67.87 \\ \hline
2 & 71.41 & 73.47 & 74.73 & 74.60 & 75.33 \\ \hline
3 & 83.87 & 85.27 & 87.40 & 85.67 & 87.47 \\ \hline
4 & 86.60 & 87.60 & 91.00 & 88.27 & 91.33 \\ \hline
5 & 91.60 & 91.13 & 92.87 & 92.07 & 93.53 \\ \hline
10 & 95.73 & 95.27 & 95.73 & 95.40 & 96.27 \\ \hline
15 & 96.53 & 96.20 & 96.27 & 96.67 & 97.00 \\ \hline
\end{tabular}
\label{tab:exp2}
\end{center}
\end{table}

\subsubsection{Results and discussion}
Table \ref{tab:exp2} shows the accuracies versus the number $N$ of training subjects.
From the table, we can see that the overall performance of CMCM was better than that of the other methods. In particular, CMCM works well when the number of training subjects $N$ is small.
For example, when $N$ is 1, CMSM and CMCM achieve an error rate of about half that for softmax.
Moreover, CMCM outperforms the subspace based methods, MSM and CMSM.
This further indicates that the convex cone based method can represent the distribution of a set of CNN features more accurately than the subspace based methods.

\section{Conclusion}\label{s:conclusion}
In this paper, we proposed a method based on the convex cone model for image-set classification, referred to as the constrained mutual convex cone method (CMCM).
We discussed a combination of the proposed method and CNN features, though our method can be applied to various types of features with non-negative constraint.

The main contributions of this paper are 1) the introduction of a convex cone model to represent a set of feature vectors compactly and accurately; 2) the definition of the geometrical similarity of two convex cones based on the angles between them, which are obtained by the alternating least squares method; 3) the proposal of a method, i.e., MCM, for classifying convex cones using the angles as the similarity index; 4) the introduction of a discriminant space that maximizes between-class variance (gaps) between convex cones and minimizes within-class variance; and 5) the proposal of the constrained MCM (CMCM), which incorporates the above projection into the MCM.

We verified the effectiveness of multiple angles
and the discriminant space which are the essence
of the proposed frameworks through two experiments.
Then, we evaluated the classification performances of the proposed methods by comparing with various types of conventional methods.
The proposed methods achieved competitive results, whether the number of training samples is large or small.

\begin{acknowledgements}
 Part of this work was supported by JSPS KAKENHI Grant Number JP16H02842.
\end{acknowledgements}

\bibliographystyle{spbasic}
\bibliography{ijcv}

\end{document}